\documentclass[10pt, a4paper]{article}
\usepackage{lrec}
\usepackage{graphicx}
\usepackage{tabularx}
\usepackage{soul}

\usepackage{epstopdf}
\usepackage[latin1]{inputenc}

\usepackage{hyperref}
\usepackage{xstring}

\usepackage{xcolor}
\usepackage{epstopdf}

\title{PoKi: A Large Dataset of Poems by Children}

\name{Will E. Hipson$^1$ and Saif M. Mohammad$^2$}

\address{$^1$Carleton University, Ottawa, Canada, $^2$National Research Council Canada\\
         williamhipson@cmail.carleton.ca,\\  saif.mohammad@nrc-cnrc.gc.ca \\}

\abstract{
Child language studies are crucial in improving our understanding of child well-being; especially in determining the factors that impact happiness, the sources of anxiety, techniques of emotion regulation, and the mechanisms to cope with stress. However, much of this research is stymied by the lack of availability of large child-written texts. 
We present a new corpus of child-written text, PoKi, which includes about 62 thousand poems written by children
from grades 1 to 12. PoKi is especially useful in studying child language because it comes with information about the age of the child authors (their grade). 
We analyze the words in PoKi along several emotion dimensions (valence, arousal, dominance) and discrete emotions (anger, fear, sadness, joy). We use non-parametric regressions to model developmental differences from early childhood to late-adolescence. Results show decreases in valence that are especially pronounced during mid-adolescence, while arousal and dominance peaked during adolescence. Gender differences in the 
developmental trajectory of emotions
are also observed. Our results support and extend the current state of emotion development research.\\ \newline 
 }

\begin{document}

\maketitleabstract

\section{Introduction} 

\noindent Adults, adolescents, and even young children use language to make sense of their feelings and to share them with others \cite{lindquist2015role}. Language is thus seen as a window into multiple aspects of emotion, such as our appraisal of objects and situations (e.g., \textit{I hate Mondays}), our emotional expressions (e.g., \textit{Off to the beach - woohoo!!}), and our emotional experiences (e.g., \textit{I feel appreciated}). 

Children 
live 
rich and varied emotional lives. Over the course of development, the way children express, experience, and communicate their emotions changes \cite{bailen2019understanding,thompson1991emotional}. 
Understanding these changes is instrumental in promoting healthy socio-emotional functioning across all stages of development. 

Children are also a vulnerable and protected section of the society. 
Thus several policies and laws are in place to protect their privacy and protect them from online manipulation/abuse. 
For example, children are prohibited to register on social media websites such as Twitter and Facebook.
One of the implications of this is that it is difficult for researchers studying child language to obtain large amounts of text written by children. 
Among the few corpora available for research are the Child Language Data Exchange System (CHILDES) \cite{macwhinney2014childes} and, in French, E-CALM \cite{doquet2013ancrages}. However, these datasets are somewhat limited in quantity and age range (e.g., CHILDES includes child-parent conversations for children ages 1-5) or quantity of text (E-CALM is limited to elementary school children).

Our research goal is to seek a better understanding of the emotional development of children from early childhood to late teens.
The first author is a senior graduate student in psychology with a focus on emotional development and the second author has a background in computational linguistics and natural language processing. Together, we present a new corpus of child-written text. 
It is a collection of nearly 62 thousand poems written by children
from grades 1 to 12. We will refer to the corpus as the 
\textit{\underline{Po}ems by \underline{Ki}ds} dataset, or \textit{PoKi} for short. The poems were already freely available online through a website by
Scholastic Corporation (a publishing, education, and media company).\footnote{http://teacher.scholastic.com/writewit/poetry/jack\_readall.asp}
We extracted the data from the website to study child language with permission from Scholastic. 
The poems came with 
the grade, first name, and last initial of the student author. The date of publishing is not available.

PoKi is especially useful in 
studying emotional development, not only becuase it includes text written by children, but also because it comes with information about the school grade of child authors, which is an adequate proxy for child age (grades 1--12 in the public school system most often correspond to 5--18 years of age). Further, even though poetry may not always reflect the inner feelings of the writer, it is a common medium for self-expression \cite{belfi2018individual}.

We analyze the words in PoKi to shed light on several research questions, including: 
\begin{itemize}
    \item Do the words that children and adolescents use in their writing reflect theories from child psychology on the developmental changes in emotions? 
    \item Do different aspects of emotions, such as valence and arousal, undergo different trajectories throughout development? 
    \item Are there gender differences in children's use of emotion words across development? 
\end{itemize}
\noindent Most past studies exploring such questions come from Psychology (see next section). They involve self-reports from a small number of children. Here, for the first time, we computationally examine tens of thousands of pieces of text (poems) written by children for emotion associations.
We also make PoKi freely available for research, with the condition that the research
be used for the benefit of children.\footnote{https://github.com/whipson/PoKi-Poems-by-Kids} We hope that this new dataset will bring fresh eyes and renewed attention to problems such as child anxiety, depression, and emotion regulation. We expressly forbid commercial use of this resource without prior consent.




\section{Related Work}

\noindent Even though emotions are central to human experience and they have been studied for centuries, we know very little about their inner workings. Two prominent models of emotions are the dimensional model and the basic emotions model.  
As per the dimensional model \cite{russell2003core},  emotions are points in a three-dimensional space of valence (positiveness--negativeness), arousal (active--passive), and dominance (powerful--weak). Here, \textit{valence} is defined similarly to the term \textit{sentiment} in the NLP work.
According to the  
basic emotions model (aka the discrete model) \cite{Ekman92,frijda1988laws,Parrot01,Plutchik80}, some emotions, such as joy, sadness, fear, etc., are more basic than others, and these emotions are each to be treated as separate categories. Each of these emotions can be felt or expressed in varying intensities. Regardless of whether one views emotions as fundamentally dimensional or discrete entities (it is beyond the scope of this paper to address this debate), we argue that language captures both of these features of emotions. 

Emotions change from one moment to the next, but also more gradually in terms of developmental time \cite{hollenstein2015time,kuppens2017emotion}. There are a multitude of co-occurring developmental factors that stimulate developmental change in the expression, experience, and understanding of emotion. These include biological/maturational changes \cite{brooks1994studying}, acquisition and implementation of different emotion regulation strategies \cite{mcrae2012development}, dynamic restructuring of interpersonal relationships \cite{de2009developmental}, and exposure and adaption to new environments and situations (e.g., school transitions) \cite{ge1994trajectories}. 

Overall, there is a trend of decreasing valence (i.e., increasing negative mood) from childhood into late adolescence, with a somewhat more pronounced drop experienced during middle adolescence (i.e., grade 10) \cite{frost2015daily,larson2002continuity,weinstein2007longitudinal}. It is less clear how the dimensions of arousal and dominance evolve over development. Regarding arousal, adolescents are thought to experience more intense emotions \cite{carstensen2000emotional,gunnar2009developmental,somerville2013teenage}, perhaps implying heightened tension and stress. As for dominance, which refers to how much control we perceive over our emotions, it is possible that dominance would decrease over time as adolescents are confronted with more emotional challenges \cite{ge1994trajectories}. There is some evidence to suggest that older adolescents express more anger compared to younger adolescents \cite{wong2018anger} and that rates of anxiety and worry increase throughout adolescence \cite{dugas2012intolerance}. 


Emotions may change differently as a function of gender. Previous studies have shown that boys have lower overall valence and their valence declines more rapidly from grades 8 to 11 compared to girls
\cite{larson2002continuity,weinstein2007longitudinal}. 
These findings are contrasted by the mental health literature which suggests that adolescent girls are more at-risk for depression and anxiety relative to boys \cite{nolen1994emergence}.

There is growing interest in working with poetry in the NLP research community.  Much of this work can be divided into two kinds: automatic poetry generation \cite{yi-etal-2018-chinese,ghazvininejad-etal-2016-generating,zhipeng-etal-2019-jiuge}
and poetry analysis \cite{mccurdy-etal-2015-rhymedesign,kao-jurafsky-2012-computational,rakshit-etal-2015-automated,fang-etal-2009-adapting}.
Much of the analysis work has looked at aspects of poems such as imagery, rhyming elements, and meter, but some work has looked at sentiment in poems as well \cite{kao-jurafsky-2012-computational,hou-frank-2015-analyzing}.
However, none of this work has examined poems written by children.
We hope that the availability of PoKi will encourage more computational work
on child language in poems.

\section{A Dataset of Poems Written by Children}
\noindent The Scholastic Corporation
hosts a website that publishes children's poems.\footnote{\resizebox{.92\columnwidth}{!}  {http://teacher.scholastic.com/writewit/poetry/jack\_readall.asp}}
It provides school-age children a platform to submit poetry which becomes openly accessible to anybody on the World Wide Web. The poems are mostly in English. The exact dates of publication are not available,
but we estimate that they are roughly from the year 2000 onwards (based on dates in the bodies of some poems). 


\indent We obtained permission from Scholastic to extract the data from the website. 
In July 2018, we scraped 62,250 poems written by children in grades 1 to 12. 
All poems came with information about the grade, first name, and last initial of the student author. 

\indent Submissions that were identical or nearly identical to a sample poem on the website  were removed. This left us with 61,330 poems.\footnote{In total, 920 poems were removed---$\sim$1.5\% of the total.} 
We refer to this dataset as \textit{Poems by Kids} or {\it PoKi}.
Original and lemmatized versions of PoKi and associated scripts are made freely available for research.\footnote{ https://github.com/whipson/PoKi-Poems-by-Kids} 

\indent Table \ref{tab:numpoems} shows the number of poems for each grade as well as the mean number of words in poems by grade (and standard deviation). Observe the high number of submissions for grades 3--7 and fewer submissions for grades 1 and 11. Observe that the poems become longer with higher grades (word count per poem  increases with grade). 
The standard deviation within poems also increases from grade 1 to 12. 

We manually inspected 120 poems (ten random poems from each grade) to determine the proportion that were written from the author's perspective. We found that 85 of the 120 poems ($\sim$71\%) were written in first person (i.e., they included tokens such as \textit{I,  my}, or \textit{we}). Although not an exhaustive check, we can infer that most poems in PoKi are likely written from the author's perspective.

Finally, we retrieved a small sample of fifty poems written by adults (from a poetry website) to serve as comparison to PoKi.\footnote{http://famouspoetsandpoems.com/top\_poems.html} Poems from this website have been analyzed by others in past CL work \cite{kao-jurafsky-2012-computational}; however, the dataset includes poems written by famous literary figures such as Maya Angelou and Walt Whitman and thus is not directly comparable to PoKi in terms of literary style and quality. Nonetheless, comparing the overall distribution of emotion words in PoKi and poems written by adults provides some initial indication of how children's poetry, in general, may differ from those of adults.

\begin{table}
{\small
\begin{small}
    	\begin{center}
	    \begin{tabular}{rrrrr}\hline
	      &    & Mean \# words & Mean \# VAD words   \\
	    	    Grade   & \#poems & per poem ($\sigma$) & per poem  ($\sigma$)\\\hline
1	&900	&37.3 (37.7)   &10.4 (10.8)\\
2	&3174	&32.1 (23.4)   &09.8 (07.0)\\
3	&6712	&35.2 (26.0)   &10.2 (07.4)\\
4	&10899	&39.3 (27.9)   &11.3 (08.0)\\
5	&11479	&44.5 (35.6)   &12.8 (09.7)\\
6	&11011	&49.6 (39.6)   &14.1 (11.4)\\
7	&7831	&59.7 (46.0)   &16.8 (12.8)\\
8	&4546	&67.6 (53.6)   &18.6 (15.1)\\
9	&1284	&91.5 (80.7)   &25.2 (22.4)\\
10	&1171	&91.8 (80.3)   &25.1 (22.3)\\
11	&667	&103.0 (104.0)   &27.8 (26.5)\\
12	&1656	&97.2 (106.0)   &27.6 (28.3)\\\hline
All	&61330	&50.3 (47.0)   &14.3 (13.0)\\
	\hline
		\end{tabular}
	\end{center}
	\end{small}
	\caption{\label{tab:numpoems} PoKi statistics by grade: Number of poems, average number of words per poem, and the average number of words from the NRC VAD Lexicon per poem. Standard deviations are shown in parentheses.}
	}
\end{table}


\section{Analyzing Emotions in PoKi}

\noindent Emotions are a key characteristic of poems. The availability of PoKi allows for the study of emotions in children's poems on a much larger scale than ever attempted before. We were specifically interested in the following questions: 
\begin{itemize}
    \item What are the average levels of emotions in poems by children? How do they compare with other sources of text such as poems written by adults?\\
    {\it Motivation:} To determine whether the words used by children in their poems are markedly different in terms of their emotion associations. 
    \item What are the correlations across different dimensions of emotions, such as valence--arousal and valence--dominance?\\
    {\it Motivation:} To better understand the kinds of emotion words used in children's poems.\\[-18pt]
    \item Are words used by boys and girls markedly different in terms of their emotion associations?\\
    {\it Motivation:} To determine impact of socio-cultural forces on the emotions expressed by girls and boys. 
    \item \textit{Our Primary Focus of Analysis:} What are the developmental trends in emotions---how do the emotion associations of words in poems change from grade 1 to grade 12?\\
    {\it Motivation:} To identify and better understand stages in children's development where they might be more or less prone to emotional distress. 
\end{itemize}
\noindent In order to address these questions, 
 we needed a method to determine the emotions associated with words, a method to identify developmental trends across the grades, and a
 way to determine
 the gender of the child author.
The three subsections below describe 
how we determine emotion associations in each of the poems using large manually created lexicons, 
how we perform non-parametric regression analysis between children's grade and emotions using Generalized Additive Models (GAMs), and
how we estimate the genders of the child authors using US census information. 




\subsection{Emotion Words in PoKi}

\noindent Tokenization of PoKi resulted in about 1.1M 
 word tokens and about 56K unique word types.
We used the NRC Valence, Arousal, and Dominance (NRC VAD) lexicon v1 \cite{vad-acl2018} and the NRC Emotion Intensity (NRC EI) lexicon v0.5 \cite{LREC18-AIL} to determine the emotion associations of the words.\footnote{http://saifmohammad.com/WebPages/nrc-vad.html\\ http://saifmohammad.com/WebPages/AffectIntensity.htm} Although we share a lemmatized version of PoKi, we opted to analyze the non-lemmatized version because the NRC VAD and EI lexica cover most morphological forms of common words. 

The NRC VAD lexicon contains about twenty thousand commonly used English words that have been scored on valence (0 = maximally unpleasant, 1 = maximally pleasant), arousal (0 = maximally calm/sluggish, 1 = maximally active/intense), and dominance (0 = maximally weak, 1 = maximally powerful). As an example, the word \textit{nice} has a valence of .93, an arousal of .44, and dominance of .65, whereas the word despair has a valence of .11, an arousal of .79, and dominance of .25. Table \ref{tab:numpoems} (last column) shows the mean number of VAD lexicon words in poems by grade.\footnote{The majority of the non-VAD terms in the poems are function words, followed by spelling errors and proper names.}

The NRC EI lexicon v0.5 contains about six thousand words from the 
NRC Emotion Lexicon (EmoLex) \cite{Mohammad13,mohammad2010emotions}
that were marked as being associated with anger, fear, sadness, and joy.\footnote{The NRC Emotion Intensity Lexicon version 1 was released recently and it includes additional English words with real-valued scores of intensity for four additional emotions: anticipation, disgust, surprise, and trust.} 
Each word comes with intensity ratings for the associated emotion---scores between 0 (lowest intensity) and 1 (highest intensity).
For instance, \textit{hate} is rated .83 on anger intensity, \textit{scare} is rated .84 on fear intensity, \textit{tragic} is rated .96 on sadness intensity, and \textit{happiness} is rated .98 on joy intensity. 

We only analyzed poems that included at least five words from the NRC VAD lexicon to ensure that each poem included a sufficient number of words for computing averages. This was the minimum cut-off and resulted in a dataset of 54,756 poems.\footnote{Experiments with other cut-offs: 0, 10, and 25 led to similar results as found with cut-off 5.} 

For each poem, we calculated the average valence, arousal, dominance, and emotion intensity scores of the words in it. 
(We also created an interactive web app where users can view PoKi poems and their VAD distributions.\footnote{https://whipson.shinyapps.io/poems\_app/}) We compared our results using this individual word approach with one that implemented simple negation handling (e.g., reversing polarity when word was preceded by a negator) and arrived at identical conclusions in terms of developmental trends and gender differences.

\subsection{Developmental Trends in Emotion}
\noindent 
We were especially interested in exploring how the emotions in PoKi change from early childhood to late adolescence.
Thus, we performed non-parametric regression analyses between children's grade and 
emotions associated with words in PoKi from grade 1 to 12 (to detect potential nonlinear trends). This would reveal, for instance, whether aspects of emotion peak or trough during certain developmental periods. 
We used Generalized Additive Models (GAMs) because they can model a smooth relationship between two variables without 
imposing strict parameter values on this relationship \cite{hastie1990exploring,wood2017generalized}.

We used the mgcv package \cite{wood2019package} to model nonlinear trends in emotion over grade.\footnote{https://cran.r-project.org/web/packages/mgcv/index.html} mgcv uses Penalized Iterative Least Squares to penalize model fit as smoothing becomes more complex. It does this by minimizing the Generalized Cross Validation (GCV) score, which is an index of model misfit that increases with respect to least squares and model complexity. 

We ran separate GAMs for each of the three dimensions: valence, arousal, and dominance, as well as anger, fear, sadness, and joy intensities. We entered grade as a predictor in all models and controlled for the linear increase in word count over time. We added gender (male vs.\@ female) in subsequent models, as a pseudo-interaction term. 
Note that GAMs do not allow for multiplicative interaction terms.\footnote{In this analysis, the gender term models a separate smooth term for each gender. To determine whether these trends differ, it is more informative to look at the model-implied trends and the 95\% CIs surrounding these estimates. 
We ran diagnostics in terms of k-index for each model to determine whether the amount of smoothing is optimal \cite{wood2017generalized}. For all models 
the k-index suggested reasonable smoothing.}

\begin{table}
{\small
\begin{small}
    	\begin{center}
	    \begin{tabular}{lrrrr}\hline
			 Gender strongly         & & &\\ 
			 associated with name           &  \#poems    &\%\\ \hline

	        female ($\geq$ 95\%)                 & 28,468    &46\\
	        male ($\geq$ 95\%)                   & 18,067    &29\\
	        neither                  & 6,294     &10\\
	        no gender information    & 8,490     &14\\
	        author name missing      & 11        &0\\\hline 
	        Total                    & 61,330    &100\\
			\hline
		\end{tabular}
	\end{center}
	\end{small}
	\caption{\label{tab:numPoemsByGender} Number of authors and poems in PoKi that are strongly associated with a gender. }
	}
\end{table}

\subsection{Determining Gender from First Name}

\noindent We used the 
baby name information from the open-access US Census to infer author gender.\footnote{https://www.ssa.gov/oact/babynames/limits.html} 
 We calculated the association of a first name with a gender as the percentage of times the name corresponds to that gender in the 2017 census data.
(We chose 2017 because of the large number of entries---27,890 first names and the number of boys and girls with each name.)
We consider a name to be strongly associated with a gender if the percentage is $\geq 95$\%.\footnote{A choice of other percentages such as 90\% or 99\% would also have been reasonable.}
If a PoKi author's first name is one of these strongly gender-associated names, then
for our experiments, we consider the author to belong to the associated gender.
This approach is not meant to be perfect, but a useful approximation in the absence of true author gender information.\footnote{We acknowledge that children and adolescents may identify their gender as non-binary, but we did not have the data to explore this. We also acknowledge that US census information is not representative of the names of children from around the world. The vast majority of the poems in PoKi are from children residing in the US, but there are submissions from other parts of the world. Chinese origin names tend not to be as strongly associated with gender as names from other parts of the world. Thus the gender analysis is mostly representative of US children.}

Similar approaches were used in the past by others to infer gender \cite{mohammad2020gender,lariviere2013bibliometrics,vanetta16,knowles2016demographer}. 
\newcite{mohammad2020gender} evaluated the above method on over 6,000 authors of Natural Language Processing papers, where the ground truth gender information was available. The method had a precision of 98.5\%.

Table \ref{tab:numPoemsByGender} shows the number of poems that were identified as written by boys, and the number of poems identified as written by girls.
Observe that the dataset has a markedly higher number of poems written by girls, than by boys. 

\begin{table}
{\small
\begin{small}
    	\begin{center}
	    \begin{tabular}{lrrrr}\hline
			\bf Emotion          & Average ($\sigma$) &  2.5\% limit & 97.5\% limit\\ \hline
		\bf PoKi \\
		VAD &&\\
			$\;\;\;$ valence     &  .63 (.11)     & .41  & .84\\
			$\;\;\;$ arousal     &  .45 (.09)     & .30  & .63\\
			$\;\;\;$ dominance   & .49 (.08)      & .34  & .65\\
		EI & &\\
			$\;\;\;$ anger       & .44 (.22)      & .00  & .83\\
			$\;\;\;$ fear        & .46 (.21)      & .09  & .84\\
			$\;\;\;$ sadness     & .42 (.20)      & .17  & .80\\
			$\;\;\;$ joy         & .49 (.18)      & .12  & .83\\[3pt]
		\bf Poems by Adults \\
		VAD &&\\
			$\;\;\;$ valence     &  .57 (.05)     & .48  & .68\\
			$\;\;\;$ arousal     &  .43 (.05)     & .35  & .53\\
			$\;\;\;$ dominance   & .50 (.07)      & .41  & .65\\
		EI & &\\
			$\;\;\;$ anger       & .48 (.19)      & .13  & .73\\
			$\;\;\;$ fear        & .46 (.18)      & .17  & .72\\
			$\;\;\;$ sadness     & .47 (.17)      & .15  & .74\\
			$\;\;\;$ joy         & .44 (.14)      & .18  & .71\\
			\hline
		\end{tabular}
	\end{center}
	\end{small}
	\caption{\label{tab:averages} Average VAD and EI scores of poems in PoKi and poems written by adults.}
	}
\end{table}

\section{Results}

\subsection{Overall VAD and Emotion Intensities}

\noindent Table \ref{tab:averages} shows the average VAD and EI scores of poems in PoKi and poems written by adults.
Although, theoretically, the average scores can range from 0 to 1, the majority of these are contained within a much narrower range. Therefore, it is useful to frame our results with respect to the score ranges containing 95\% percent of poems. The upper (97.5\%) and lower limits (2.5\%) of the scores are listed in Table \ref{tab:averages}.
The mean valence is significantly higher in PoKi than in the poems by adults ($p < .001$); the differences in mean arousal and mean dominance are small.


\begin{table}
{\small
\begin{small}
    	\begin{center}
	    \begin{tabular}{lrrrr}\hline
			                                &  PoKi     &  Adults   &  VAD Lexicon\\ \hline
			$\;\;\;$ valence--arousal       & $-.06$   &$-.07$     & $-.27$\\
			$\;\;\;$ arousal--dominance     &    $.52$  &$.37$    & $.30$\\
			$\;\;\;$ dominance--valence     &     $.47$  &$.31$   & $.49$\\
			\hline
		\end{tabular}
	\end{center}
	\end{small}
	\vspace*{-3mm}
	\caption{\label{tab:corr} V--A, A--D, and D--V correlations in PoKi, Adult Poems, and in the NRC VAD Lexicon. }
	}
\end{table}

\subsection{V--A, A--D, and D--V Correlations}

\noindent Although valence, arousal, and dominance are theorized to be orthogonal, we explored correlations among these scores in the current dataset. Table \ref{tab:corr} shows the V--A, A--D, and D--V correlations for the words in PoKi, poems by adults, and also for all the words in the NRC VAD Lexicon.
Valence was significantly negatively associated with arousal (r = -.06, $p < .001$), although the effect was small, suggesting minimal collinearity. However, correlations with dominance (both A--D and D--V) were much stronger and significant ($p < .001$).
These correlations are consistent with previous findings in VAD Lexicon \cite{warriner2013norms,vad-acl2018}, and perhaps call into question the orthogonality of three dimensions which, indeed, remains a contentious issue in emotion theory \cite{fontaine2007world}.


\begin{table}[t]
{\small
\begin{small}
    	\begin{center}
	    \begin{tabular}{lrrrrrr}\hline
			            & \multicolumn{2}{c}{Valence}      & \multicolumn{2}{c}{Arousal}    & \multicolumn{2}{c}{Dominance}\\
			            & mean  &$\sigma$ & mean  &$\sigma$ & mean  &$\sigma$\\\hline
			females (F)    & .636 & .11  & .456  & .09  & .490 & .08\\
			males   (M)    & .613 & .11  & .443  & .08  & .490 & .08\\
			\hline
		\end{tabular}
	\end{center}
	\end{small}
	\caption{\label{tab:genderVAD} Mean valence, arousal, and dominance in poems written by
male and female child authors. The F and M differences in valence and arousal are statistically significant.}
	}
\end{table}

\subsection{Gender Differences in Emotions}

\begin{figure*}[t!]
 \begin{center}
 	\includegraphics[width=2\columnwidth]{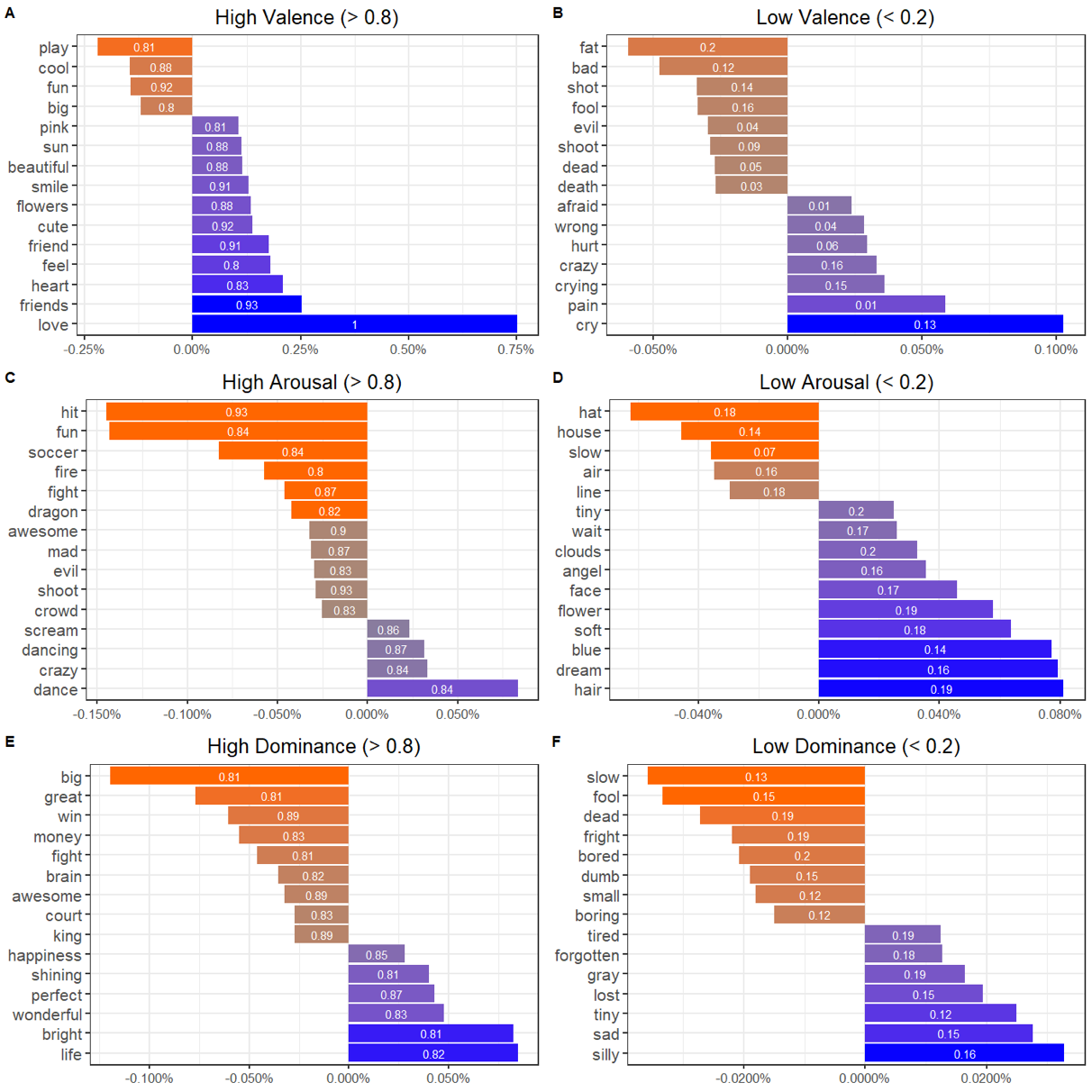}
 	\vspace*{3mm}
 	\caption{Gender differences in word occurrences for High Valence (A), Low Valence (B), High Arousal (C), Low Arousal (D), High Dominance (E), and Low Dominance (F) words. X-axis shows average percent difference in word usage between boys and girls with positive percentage reflecting more usage by girls. Y-axis shows 15 words that differ the most in usage between boys and girls on the respective dimension. Values inside the bars are the VAD scores for the words on the respective dimensions. }
 	\label{fig:Fig2}
 \end{center}
 \vspace*{-3mm}
 \end{figure*}

\noindent Table \ref{tab:genderVAD} shows the mean valence, arousal, and dominance in poems written by
male child authors and female child authors. It also includes the standard deviations.
Observe that, on average, poems written by females 
have higher valence (are more positive) compared to those by males.
A notable driver of this difference is the word \textit{love} (which has a high valence score) which makes up 1.94\% of words in poems written by females and only 1.19\% in poems written by males. On average, poems written by males 
have higher arousal compared to those by females. 
The word \textit{fun} (which has a high arousal score) makes up .67\% of words in poems written by males and only .52\% in poems written by females.
The difference in average dominance scores of poems by females and males is not statistically significant (\textit{t}(1, 41558) = .762, $p = .462$). 
Figure \ref{fig:Fig2} shows which NRC VAD words had the greatest difference in occurrence between males and females.

Table \ref{tab:genderEI} shows the mean anger, fear, joy, and sadness intensities in poems written by
male child authors and female child authors. On average, the poems written by males have a significantly higher mean intensities for all three negative emotions (anger, fear, and sadness) than poems by females. On average, the poems written by females have a significantly higher mean intensities for joy.



\subsubsection{Trends in VAD by Grade}
\begin{figure*}[t]
 \begin{center}
 	\includegraphics[width=2\columnwidth]{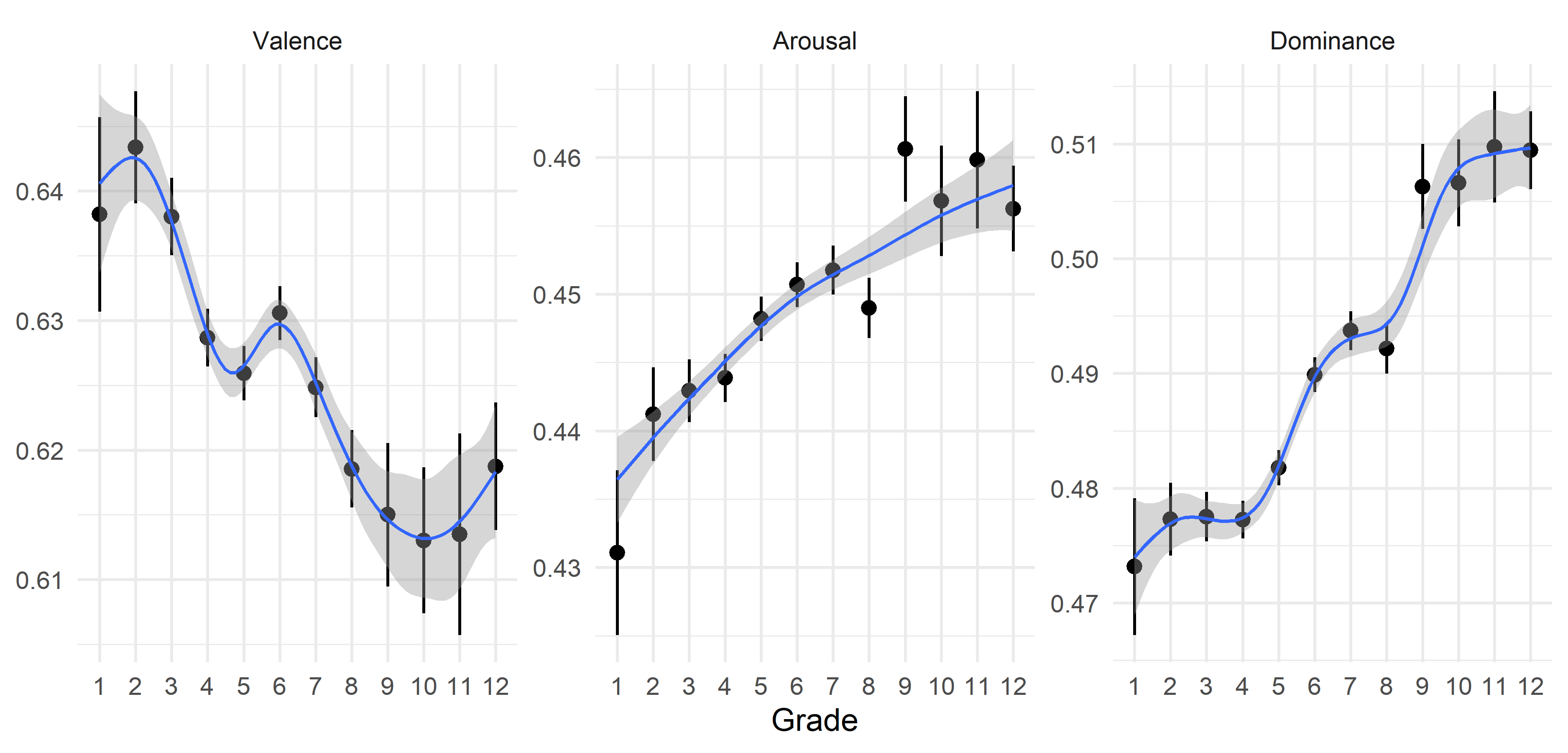}
 	 \vspace*{-3mm}
 	\caption{Mean VAD scores by grade (dots) as well as model-implied trends. 
 	The grey band represents 95\% CI around the smooth fit (GAM).}
 	\label{fig:Fig3}
 \end{center}
 \vspace*{-3mm}
 \end{figure*}

\noindent Figure \ref{fig:Fig3} shows mean VAD scores by grade (dots) as well as model-implied trends by grade (the curved line connecting the dots). The grey band represents 95\% confidence intervals (CI) around the smooth fit (GAM).\\[-8pt]

\noindent {\bf Valence:} Observe that mean valence decreased by about 4\% from grade 1 to grade 12. Given that 95\% of the scores for valence in this sample lie between .41 and .84, this represents a roughly 10\% decrease within the range of valence scores. Model-implied trends show a steeper decrease into grades 9 and 10, after which valence levels off. 
The relation between grade and valence was statistically significant.\\[-8pt]

\noindent {\bf Arousal:} Results for arousal showed an overall upward trend. The overall increase in arousal was also 4\%, which translates roughly into a 12\% increase within the typical range of arousal scores. \\[-8pt] 

\noindent {\bf Dominance:} Mean dominance over grade showed a similar pattern to that of arousal, increasing in an upward, curvilinear fashion, and peaking around grade 9--12. Overall, the increase amounted to roughly 15\% increase in dominance within the typical range in dominance scores. 

\subsubsection{Differences Across Genders in VAD by Grade}

\begin{table}[t]
{\small
\begin{small}
    	\begin{center}
	    \begin{tabular}{lrrrrrrrr}\hline
			            & \multicolumn{2}{c}{anger}      & \multicolumn{2}{c}{fear}    & \multicolumn{2}{c}{joy} & \multicolumn{2}{c}{sadness}\\
			            & mean  &$\sigma$ & mean  &$\sigma$ & mean  &$\sigma$. & mean  &$\sigma$\\\hline
			F     & .433 & .22  & .453  & .21  & .493 & .17 & .409 & .20\\
			M       & .462 & .22  & .473  & .20  & .465 & .18 & .426 & .20\\
			\hline
		\end{tabular}
	\end{center}
	\end{small}
	\caption{\label{tab:genderEI} Mean anger, fear, joy, and sadness intensities in poems written by
male child authors and female child authors. All F and M differences are statistically significant.}
	}
\end{table}

\begin{figure*}[t]
 \begin{center}
 	\includegraphics[width=2\columnwidth]{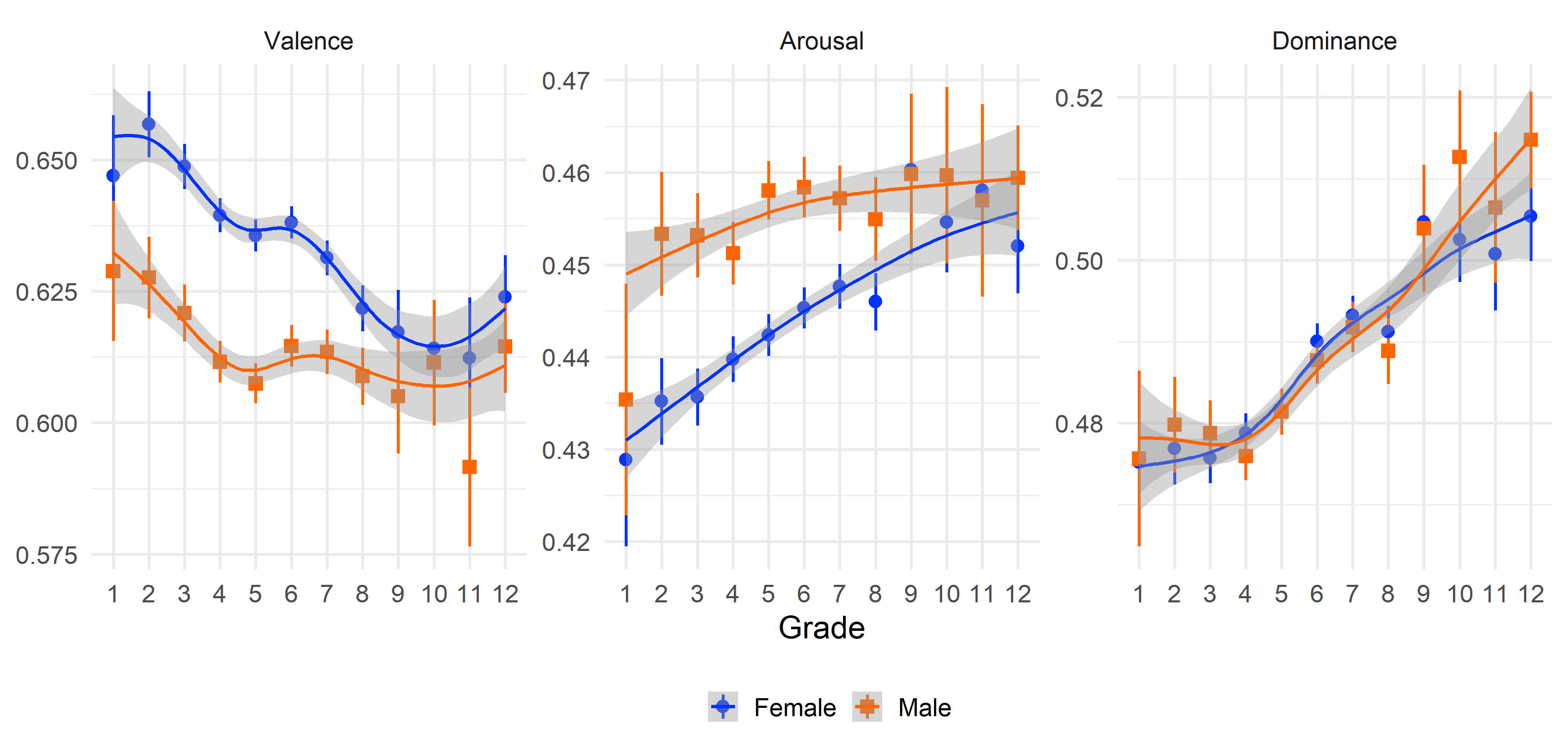}
 	 \vspace*{-3mm}
 	\caption{Mean VAD scores by grade and gender, as well as model-implied trends. 
 	Notes: Grey area represents 95\% CI around the smooth fit (GAM). Dots represent average value per grade and its 95\% CI.}
 	\label{fig:Fig4}
 \end{center}
 \vspace*{-3mm}
 \end{figure*}

\noindent Figure \ref{fig:Fig4} shows mean VAD scores by grade 
as well as model-implied trends.
The grey band represents 95\% CI around the smooth fit (GAM). \\[-8pt] 

\noindent {\bf Valence:} Observe that the model for females shows a downward sloping curvilinear relation over grade, whereas for males, the association is more linear and less steep relative to females. \\[-8pt]

\noindent {\bf Arousal:} For females, arousal increased at a steeper rate, whereas for males it is more gradual. \\[-8pt] 

\noindent {\bf Dominance:} The trend in dominance over grade did not appear to differ between males and females.

\subsection{Trends in Emotion Intensities by Grade}
\begin{figure*}[t]
 \begin{center}
 	\includegraphics[width=1.2\columnwidth]{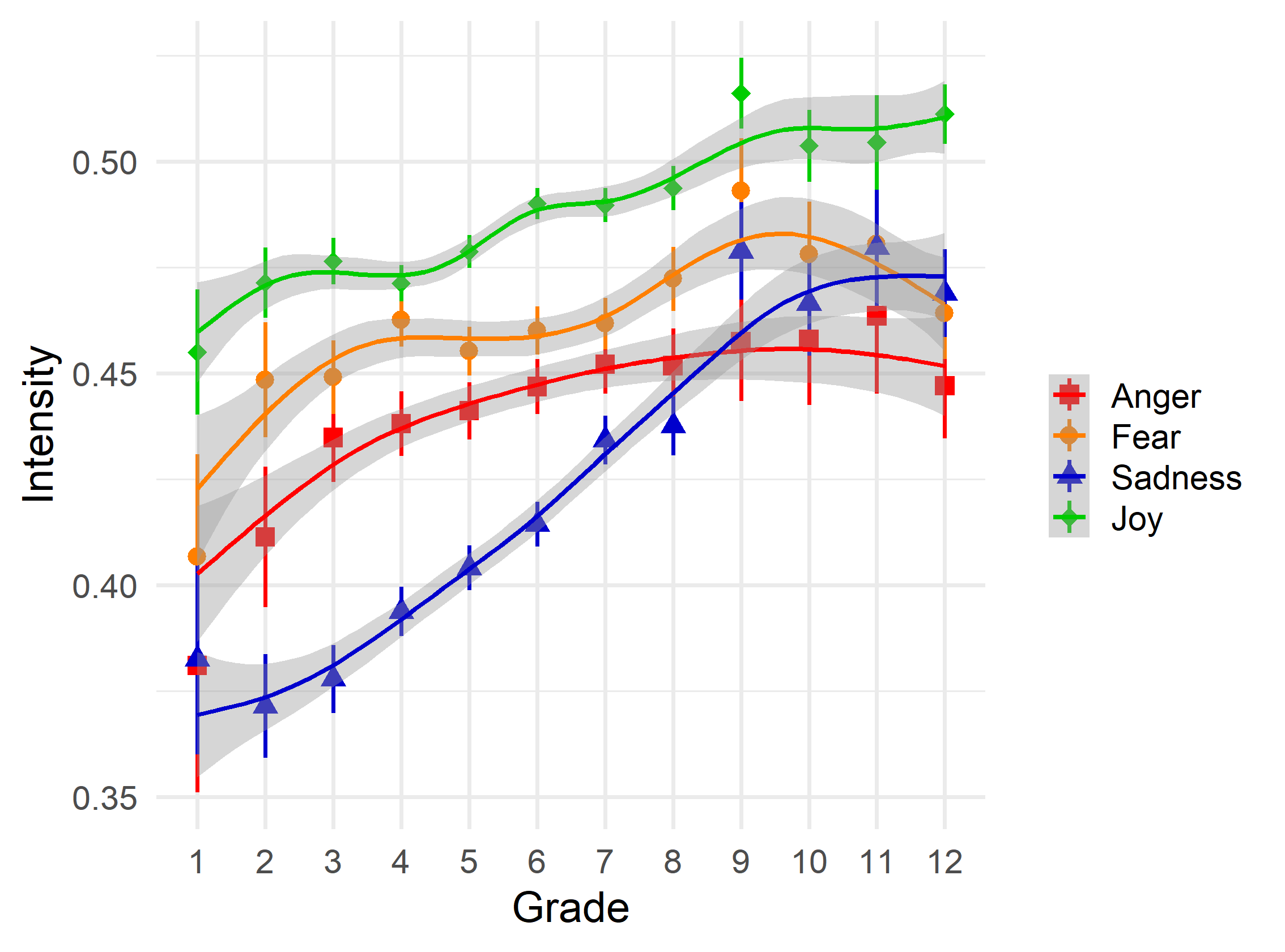}
 	 	 \vspace*{-3mm}
 	\caption{Mean  anger, fear, sadness, and joy by grade and model-implied trends.}
 	\label{fig:Fig5}
 \end{center}
\vspace*{-6mm}
 \end{figure*}

\begin{figure*}[t]
 \begin{center}
 	\includegraphics[width=1.4\columnwidth]{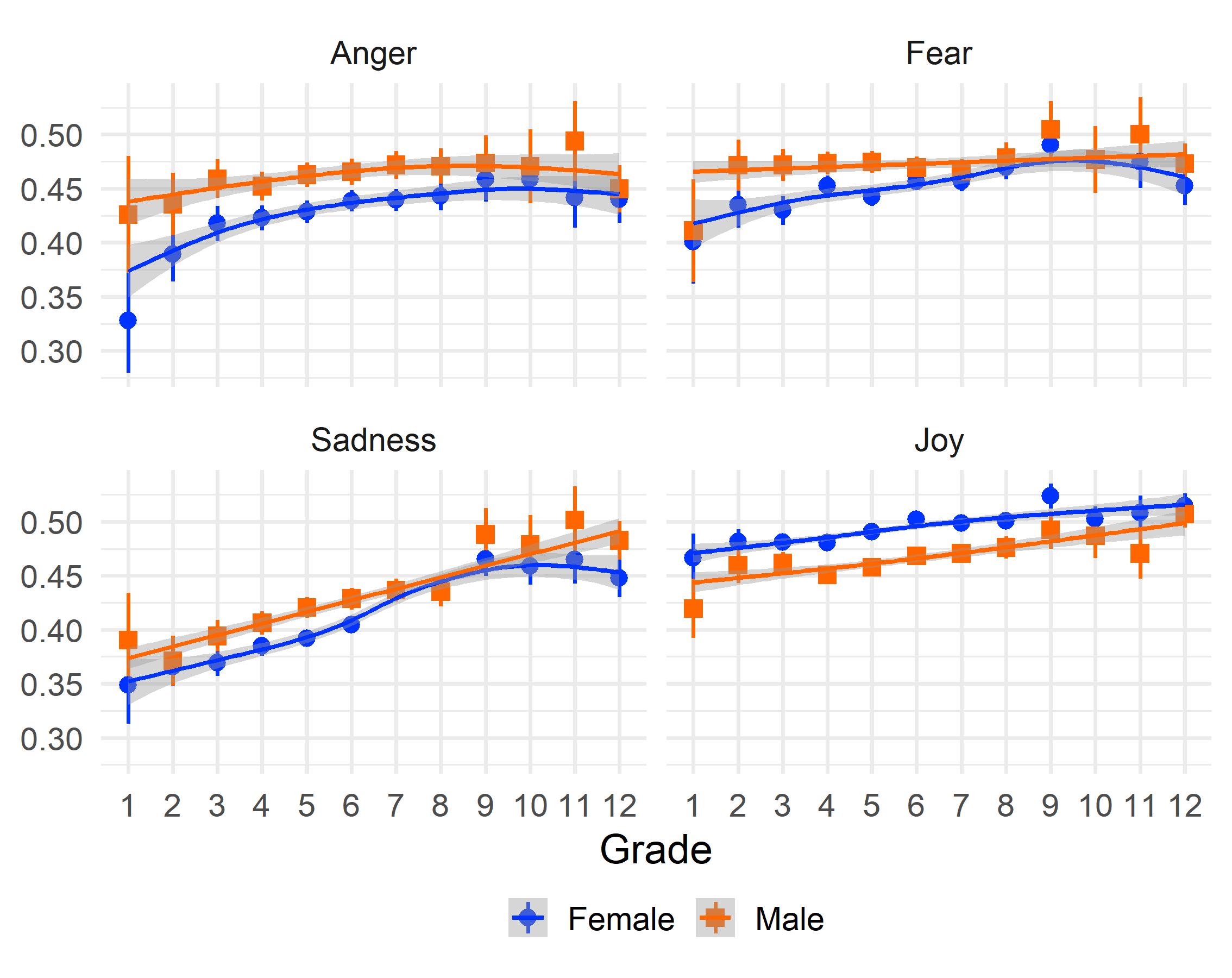}
 	 	 \vspace*{-3mm}
 	\caption{Mean  anger, fear, sadness, and joy by grade and gender, as well as model-implied trends.}
 	\label{fig:Fig6}
 \end{center}
\vspace*{-3mm}
 \end{figure*}

\noindent Figure \ref{fig:Fig5} shows the mean  anger, fear, sadness, and joy by grade and model-implied trends. The grey area represents 95\% CI around the smooth function. 
Figure \ref{fig:Fig6} shows means and model trends by gender.\\[-10pt]

\noindent \textbf{Anger Intensity:} Anger increased from grade 1 to 3 by about 7\% of the typical range in anger and leveled off thereafter. For the most part, the trends appeared similar between both genders, although there was perhaps a stronger increase in anger among females between grades 1 and 2.\\[-2pt]

\noindent \textbf{Fear Intensity:} Results for fear were similar to those of anger. Fear increased more strongly from grade 1 to grade 3 by about 4\% and then increased again slightly around grade 10. In this case, however, there did not appear to be different trends for males and females.\\[-8pt]

\noindent \textbf{Sadness Intensity:}
Sadness showed the strongest increase over the course of development, increasing by about 12\%, or 19\% within the typical range, and peaking in grade 11. The model predicting sadness by grade also fit relatively better than the other emotions. 
In terms of gender differences, the pattern of increase for females appeared more curvilinear during grades 6 to 8, whereas for males it was strictly linear throughout.\\[-8pt]

\noindent \textbf{Joy Intensity:}	Joy showed a pattern of sinusoidal (wave-like) increase, peaking in grade 9. Joy was also consistently higher than any of the negative emotions. 
The trends did not appear to vary by gender.

\section{Discussion}

\noindent \textit{Trends in VAD:} Our analyses into the emotions associated with words in children's poems found interpretable effect sizes that are largely consistent with previous research in Psychology. For valence, our results are consistent with previous research using self-reports \cite{frost2015daily,larson2002continuity,simmons1987impact,weinstein2007longitudinal}. Specifically, we found that average poem valence decreased over time, most precipitously during grades 6-9 and reaching its lowest at grade 11. The trends further indicate that valence begins to rebound in grade 12, which may reflect a readjustment of positive mood in late-adolescence/emerging adulthood. Broadly speaking, the current findings are consistent with research suggesting that the early- to middle-adolescence transition is an emotionally challenging period of development \cite{arnett1999adolescent}. However, the magnitude of this decrease does not imply that early-mid adolescence is a particularly calamitous developmental period \cite{hollenstein2013beyond}. Overall though, our results suggest that developmental differences in emotion language correspond with previously identified trends in self-reported mood.

Although much of the empirical work on emotional development has focused on valence, there is reason to suspect developmental changes in arousal and dominance. Our results showed that average poem arousal increased throughout development. This is consistent with the view of adolescence as a time when emotions are more charged and intense \cite{carstensen2000emotional,gunnar2009developmental,somerville2013teenage}. For example, compared to children and adults, adolescents react more visibly to images of happy and fearful faces, suggesting a heightened sensitivity to emotional arousal \cite{dreyfuss2014teens,somerville2011frontostriatal}. 

As for dominance, the patterns mirrored those of arousal. 
Dominance is somewhat difficult to interpret in the context of emotion language. Interpreting it at face value, we might conclude that the results reflect increased capabilities in emotion regulation (i.e., being more in control of one's emotions) \cite{zimmermann2014emotion}. However, we are hesitant to make this conclusion because individual words likely have poor correspondence with emotion regulation, which involves complex processes. In contrast, the similarity between arousal and dominance in this analysis suggests to us that these may be tapping aspects of the same construct (e.g., emotional salience).

\noindent {\it Gender differences in VAD 
over development:} Although the overall higher valence of female poems is consistent with previous linguistic analyses \cite{newman2008gender}, contrary to our hypothesis, there appeared to be a stronger decrease in valence in poems written by females. This finding is inconsistent with previous research on developmental changes in negative mood, which is more salient among boys \cite{weinstein2007longitudinal}. One way of looking at this is through girls' superior emotion vocabulary and emotion comprehension, which may allow girls to incorporate more positively valenced terms in their writing at an earlier age \cite{bosacki2004preschoolers}. This may then result in a stronger decrease in girl's valence expression as they experience more negative emotions in adolescence \cite{chaplin2013gender}. 

There is marked difference in the trend for arousal across genders. It is interesting to note that from as early as grade 2, the amount of arousal in male poems is markedly higher than that in female poems. This difference gradually decreases with the linear increase in arousal in female poems with grade. 
 

Finally, the trends for dominance appear to be identical across gender. This is also surprising as one might expect that gender norms would encourage boys to express greater dominance (e.g., use more powerful and active words) at any earlier age compared to girls.

\noindent {\it Trends in anger, fear, sadness, and joy intensities:} We found that anger, fear, and sadness intensity all increased from grade 1 to 12. Sadness intensity showed a particularly strong increase during this period, which dovetails with the notion that adolescents report more negative, depressed mood \cite{holsen2000stability}. In contrast, joy increased throughout development. We suspect that this is because high joy terms such as \textit{love} and \textit{heart} are more common among adolescent poems. Overall, analysis of discrete emotion intensities shows some patterns distinct from global trends in valence, arousal, and dominance, but gender differences are probably better explained by overall mean differences.

Taken together, our analysis of emotion language in a large corpus of poetry corroborate and extend what was previously known about emotion development. Our results highlight important differences in the language used to denote varieties of emotional experience, but also support a general trend of increasing emotion intensity. Our research does not go so far as to explain what drives these differences, but we can turn to previous research for hints. Most compelling is that our results correspond with developmental trajectories in mood \cite{frost2015daily,larson2002continuity,weinstein2007longitudinal}. 
However, we also know that emotion language shifts over development from a focus on subjective feelings to external features \cite{o2004developmental}. As well, vocabulary development allows older children to use more abstract, figurative language that may contain affect-laden terms \cite{demorest1983telling}. 

\subsection{Limitations and Ethical Considerations}
\noindent We used a simple approach that averaged emotion associations across the words in each poem, rather than considering word order or semantic structure. This point is especially important for an analysis of poetry, which relies heavily on figurative language. For example, one poem in this sample included the words monster, ghost, witch, and scary. Our algorithm scored this poem as high in fear, but the poem is clearly about Halloween and thus seemed more lighthearted. 
We should also question to what extent counting emotion words is reflective of children's emotional states. 

\newcite{kahneman2005living} propose that our experience of emotions is different from how we remember our emotions, so even if children were recounting past experiences in their poems, the affective content might differ from how the experience actually felt. Although our findings generally fit with theorized patterns of emotional development, we cannot say for certain whether the current results reflect changes in felt emotion, emotion vocabulary or, more broadly, changes in the ability to use abstract language. Therefore, we are more inclined to cautiously interpret our results as reflecting developmental trends in the distribution of emotion words. 

There are potential limitations with the sample that we used. Although we argue that poetry is a medium for self-expression, we acknowledge that poetry plays with abstract ideas and emotions, which makes it perhaps not the best source for identifying felt emotion. 
Finally, it is difficult to know children's motivations for submitting these poems and how their motivations may differ at different ages. The poems were hosted on an educational resource platform, suggesting that children may have written and submitted the poems for class assignments. This may have something to do with the decrease in poems submitted by older adolescents. It would be interesting to compare the current results with a different sample of child and adolescent writing. 

Despite clear benefits to studying child language, such as improving our understanding of child anxiety and well being, NLP research on child language can be abused, for example, to manipulate children's emotions online.\footnote{We acknowledge that PoKi contains the author's first name and thus is not strictly considered as anonymized. However, we view it as highly unlikely that authors can be identified based on first name only. Moreover, the poems---including author name---are all freely accessible via Scholastic.org and we obtained permissions from the domain host to analyze and share this data.} Thus the terms of use of this resource require that it be used only for the benefit of children. We also expressly forbid commercial use of the resource without prior consent. 

\section{Conclusions and Future Directions}
\noindent We presented the \textit{Poems by Kids Corpus (PoKi)}---a corpus of about 62 thousand poems written by children.
PoKi is especially useful in studying child language because it comes with information about the age of the child authors (their grade). 
We analyzed the emotions associated with the words in PoKi to shed light on several research questions. 
 We used nonparametric models to describe nonlinear trajectories over the course of development. We were also able to infer author gender for most poems to see how developmental differences vary between males and females. 
 We showed that the words that children and adolescents use in their poetry reflect important developmental changes in emotion expression. 

The Poems by Kids Corpus has broad potential for future research and application. For instance, focused approaches on developmental differences in text complexity and semantic structure could inform the creation of reading guidelines in educational contexts \cite{tortorelli2019beyond}.
PoKi can be used to understand the factors that drive or trigger  emotional experiences in children.
It can be used to train machine learning models to generate poetry as children do. 
The dataset also has applications in developing conversational agents geared towards children and pedagogy \cite{tamayo2017designing,perez2013exploratory,mower2011rachel}.
We make the corpus freely available for research.


\section{Bibliographical References}
\label{main:ref}

\bibliographystyle{lrec}
\bibliography{poems,anthology}


\end{document}